\documentclass[runningheads]{llncs}

\usepackage{eccv}

\usepackage{eccvabbrv}

\usepackage{graphicx}
\usepackage{booktabs}

\usepackage[accsupp]{axessibility}  

\usepackage{hyperref}

\usepackage{orcidlink}

\usepackage{newtxtext,newtxmath}
\usepackage{multirow}
\usepackage{tabularx}
\usepackage{mathtools}
\usepackage{graphicx}
\usepackage{subcaption}

\usepackage{amsmath}
\usepackage{amssymb}
\usepackage{booktabs}
\usepackage[T1]{fontenc}

\usepackage{bbding}

\begin{document}

\title{Multi-Scale Representation Alignment for Visual Autoregressive Modeling with Mixture of Experts}

\titlerunning{MEPA}

\author{Nuoyan Zhou\inst{1,2}$^{\dagger}$\orcidlink{0009-0009-2424-4388} \and
Zhijun Tu\inst{2}\orcidlink{0000-0001-8740-7927} \and
Lei Yu\inst{2}\and
Kun Cheng\inst{1}\orcidlink{0000-0002-6399-8424} \and
Jie Hu\inst{2}$^{*}$ \and
Nannan Wang\inst{1}$^{*}$\orcidlink{0000-0003-1435-489X}  \and
Xinghao Chen\inst{2}\orcidlink{0000-0002-2102-8235}
}

\footnotetext{\hspace*{-1.2em}$^{\dagger}$ This work was done during an internship at Huawei Technologies Co., Ltd.}
\footnotetext{\hspace*{-1.2em}$^{*}$ Corresponding authors}

\authorrunning{N. Zhou et al.}

\institute{State Key Laboratory of Integrated Services Networks, Xidian University, Xi'an, China 
nuoyanzhou@stu.xidian.edu.cn, kuncheng.xidian@gmail.com, nnwang@xidian.edu.cn \and
Huawei Technologies Co., Ltd. \\\{zhijun.tu, yulei96, hujie23, xinghao.chen\}@huawei.com
}

\maketitle

\begin{abstract}
  Visual AutoRegressive modeling (VAR) has pioneered a coarse-to-fine multi-scale autoregressive generative paradigm, demonstrating strong capabilities in image generation. However, VAR still suffers from inherent deficiencies in multi-scale representation learning. Specifically, lower scales primarily capture global semantics, while higher scales focus on fine-grained details. Employing a shared architecture across scales induces optimization conflicts. Moreover, due to the causal autoregressive process, inaccurate semantics at early scales can propagate and significantly degrade the final output. To address these issues, we introduce a scale-aware token-routed Mixture of Experts (MoE) architecture, allowing scale-adaptive expert selection, thereby facilitating decoupled representation learning across scales. In addition, we enhance semantic modeling at early scales by incorporating external self-supervised features. Unlike naive alignment, we analyse and design a residual feature aggregation scheme tailored to the VAR paradigm. Extensive experiments show that our method significantly improves both training efficiency and generation quality. On the ImageNet 256×256 benchmark, our model achieves a superior FID compared to the dense baseline while requiring only half of the default training epochs and a smaller parameter budget, with a merely marginal increase in training cost. Moreover, the performance gap further widens with larger training epochs.
  
\end{abstract}

\section{Introduction}
\label{sec:intro}

Autoregressive (AR) models, empowered by their remarkable scaling laws~\cite{Scaling-laws-nlp, Scaling-laws-cv}, have achieved tremendous success in natural language processing (NLP)~\cite{gpt4,gemini,llama,bert}. This success has inspired their adaptation to the computer vision~\cite{PixelCNN,vqgan,emu3,show-o,vargpt}, where researchers aim not only to replicate the breakthroughs witnessed in NLP, but also to explore AR modeling as a unified paradigm for both image generation and understanding tasks~\cite{chameleon, emu3}. A typical AR paradigm for image generation involves two stages: it first discretizes images into tokens via a tokenizer~\cite{vqvae,vqgan, VAR, maskbit,titok,softvq,mgvq} and then utilizes an AR network to generate samples in a sequential manner. Built on this foundational paradigm, recent works~\cite{llamaGen, randar, VAR, MAR, rar, xAR, PAR, spectralar} have made great progress in image generation, demonstrating comparable performance to diffusion models.

\begin{figure*}[!htbp]
  \centering
  \includegraphics[width=\linewidth]{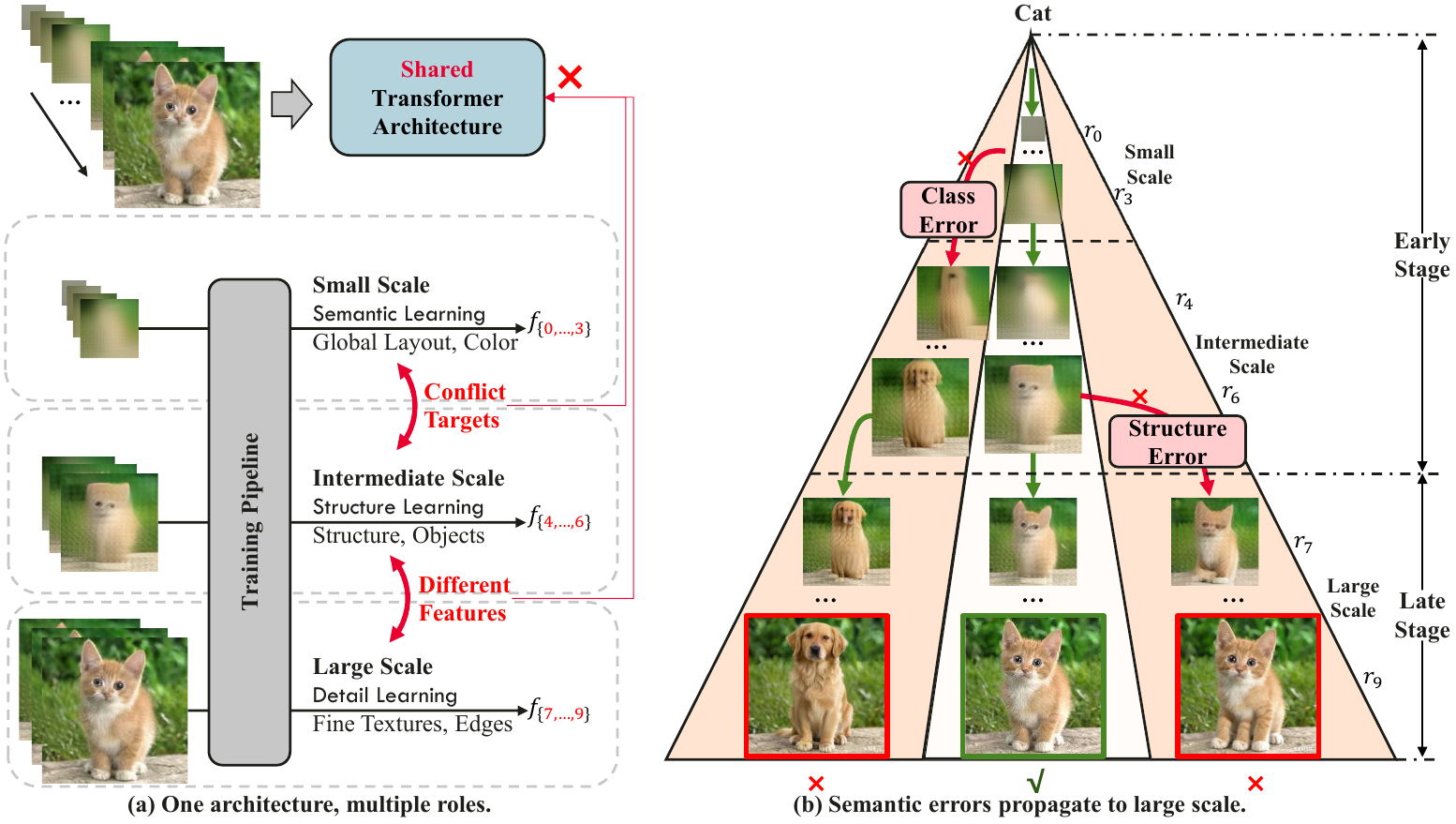}
  \caption{\textbf{Empirical analysis of representation space and expert load}. (a) VAR learns different features across scales with a shared architecture, resulting in conflicts of optimization objectives. (b) Semantic errors at small and intermediate scales can propagate across scales and induce unpleasing results. Correct semantics at early stage are critical for predictions at later stage.}
  \label{motivation}
\end{figure*}

Specifically, Visual AutoRegressive modeling (VAR) ~\cite{VAR} stands out among AR methods for its high-quality and fast image generation. It has pioneered a coarse-to-fine autoregressive modeling via next scale prediction, which decomposes the image data into multi-scale residual representations and builds causality across scales. This design preserves the 2D structure of images and supports efficient parallel decoding, leading to excellent performances in both generation quality and inference speed. However, it suffers from difficulties in multi-scale representation learning. In this paper, we argue that a Mixture of Experts (MoE)~\cite{MoE} architecture and semantic guidance in the representation space can substantially mitigate these issues.

We start by analysing the feature distribution of VAR representations. As shown in Fig.\ref{motivation}a, different scales focus on different learning targets and generate distinct feature spaces, which requires the model to play roles with multiple capabilities across scales. Since all the scales share a single Transformer architecture, these differences induce conflicts in optimization and difficulties in multi-scale representation learning. Besides, due to causal nature across scales, predictions at higher scale depends heavily on the feature map produced at earlier scales. As shown in Fig.\ref{motivation}b, inaccurate semantics at early scales can propagate and adversely affect later scales, leading to unpleasing outputs.

To address these challenges, we propose \underline{m}ulti-scale r\underline{ep}resentation \underline{a}lignment (MEPA), a token-routed Mixture of Experts (MoE) framework for guiding efficient representation learning with semantically-enriched visual representations. First, we introduce a MoE architecture that adaptively decouples model capacity across scales, thereby implicitly reducing the adverse effect of conflicting optimization and promoting specialized feature learning. Our approach evolves from a scale-based routing scheme to a size-aware token routing scheme, due to the latter preventing routing homogenization within the same scale caused by the former and improving load balancing of expert training(see Sec.~\ref{moe}). Next, we incorporate self-supervised representations to guide the semantic information at early stage in VAR models, mitigating the propagation of semantic errors throughout the autoregressive generation process. However, there exists an inherent feature gap between VAR models and self-supervised models, which prevents direct alignment~\cite{repa, VAVAE}. Thus, we conduct a series of empirical analyses on transforming VAR features to bridge the representation gap and inject accurate semantic information (see Sec.~\ref{alignment}). To the best of our knowledge, this is the first systematic study of representation alignment in the VAR paradigm. Our experiments show that MEPA significantly improves both training efficiency and generation quality. It achieves superior performances compared to the baseline with a half of the default training epochs and a smaller parameter budget. Moreover, the performance advantages further widens with larger epochs. We highlight our main contributions below:
\begin{itemize}
  \item We highlight the challenges of multi-scale representation learning in VAR, systematically analysing the inherent difficulties caused by the shared architecture and semantic errors propagation.
  \item We propose MEPA, a scale-aware MoE framework that promotes decoupled representation learning across scales and strengthens semantic information.
  \item MEPA achieves impressive performances on both training efficiency and generation quality. It reaches up to a 2× training speedup over the VAR training process.
\end{itemize}

\section{Related Works}
\label{sec:Related}

\textbf{Autoregressive Model in Image Generation.} The autoregressive (AR) methods mostly adopt a two-stage paradigm, which discretizes images into quantized tokens by a pretrained tokenizer~\cite{vqvae,vqgan,VAR,maskbit,titok,mgvq} and then generate samples in a sequential manner. Built on this foundational paradigm, recent works~\cite{llamaGen,randar,VAR,MAR,rar,xAR,PAR,spectralar} have made great progress in the performance of image generation. The basic unit generated at a single AR step can take various forms (such as token~\cite{llamaGen,rar,randar}, scale~\cite{VAR,MVAR,infinity,flexvar,spectralar}, patch~\cite{xAR,npp}, frequency~\cite{FAR,nfig,SIT}, etc~\cite{PAR,MAR,SAR,iar}), enabling the AR model to simultaneously improve the quality of generated images and the efficiency of parallel inference. Besides, there are numerous efforts in exploring high-efficiency schemes of visual generation models. A few works~\cite{titok,flextok,softvq} focus on reduce compressed token numbers of a image. \cite{fastvar} propose a pivotal token selection strategy to generate only pivotal tokens for fast inference. \cite{MVAR} apply a scale-Markov trajectory to cut off redundant KV cache. \cite{CoDE} use a large VAR model to generate the tokens at small scales and a small VAR model to efficiently predict the remaining tokens. However, few studies have focused on the training efficiency of AR models, which is precisely what our work aims to emphasize and address.

\noindent\textbf{Self-supervised Representation Learning.} Recent advances in self-supervised representation learning can be broadly categorized into contrastive learning and masked image reconstruction. The methods trained by contrastive learning, such as including DINO~\cite{dinov2}, SimCLR~\cite{simclr}, MoCo\cite{moco}, and BYOL~\cite{byol}, learn informative features without explicitly modeling the data distribution, producing highly separable representations that excel in classification, retrieval, and dense prediction, but often discard fine grained generative information, limiting their utility for synthesis or reconstruction. The approaches trained by masked image reconstruction, including VAE~\cite{vae}, masked autoencoders (MAE)~\cite{mae}, masked image modeling (MIM)~\cite{simmim}, diffusion models~\cite{ddpm}, and autoregressive (AR) models~\cite{bert} capture rich contextual and perceptual information by reconstructing inputs or modeling the underlying data distribution, providing embeddings beneficial for downstream tasks; however, they are computationally intensive and may produce representations that are less discriminative than contrastive methods. The success of~\cite{repa, VAVAE,imagefolder} has demonstrated that generation tasks benefit from pretrained self-supervised representations, which inspires us to transfer it to the AR paradigm for the first time.

\section{Preliminaries}
\label{Preliminaries}
VAR employs next-scale prediction to generate multiple tokens simultaneously. Its tokenizer first compresses the image $x$ into a feature map $f_0 \in \mathbb{R}^{H \times W \times c}$ and decomposes it as a set of multi-scale residual features $\{f_1, f_2, \ldots, f_K\}$. Specifically, it quantizes the downsampled feature into tokens $\{r_1, r_2, \ldots, r_K\}$ in ascending order of scale, and then obtains the next-scale feature by subtracting the quantized feature from the current-scale feature. The process can be formulated as
\begin{align}
    Q(z) &= \arg \min_{v_i \in V} {\lVert v_i - z \rVert}, \\
    r_k &= Q(Down(f_{k-1}, {sz}_k)),\\
    f_k &= f_{k-1} - \sum_{i=1}^{k-1} Up(Res_i(LU(r_i)), {sz}_K),
\end{align}
where $Q(\cdot)$ denotes the quantization operation, $V$ denotes the codebook, $v_i$ denotes the $i$th feature in the codebook, $Down(\cdot)$ and $Up(\cdot)$ denotes the down and up sample operation, $Res_k(\cdot)$ denotes the ResNet block for the $k$th scale, ${sz}_k$ denotes the feature size $(H_k, W_k)$ at the $k$th scale, $LU(\cdot)$ means retrieving features from the codebook based on indices. VAR builds a causal relationship upon multi-scale features with conditions $c$:
\begin{equation}
    p(r_1,r_2,\ldots,r_K) = \sum_{k=1}^{K}p(r_k|r_1',r_2',\ldots,r_{k-1}', c), 
\end{equation}
where $r_i' = Down( \sum_{j=1}^{i} Up(Res_j(LU(r_j)), {sz}_{K}), {sz}_{i+1})$ means merging all preceding residual features as an accumulated input. In the training process, all scales except the last one are concatenated along the sequence dimension to predict the probability distribution. During the inference, VAR can generate tokens in parallel from small to large scales in a autoregressive method. The training loss can be formulated as
\begin{equation}
\label{celoss}
\mathcal{L}_{\mathrm{CE}}
= -\mathbb{E}_{x}\!\left[\, CE\big(p(r_1, r_2, \ldots, r_K);\; r_1, r_2, \ldots, r_K\big)\right],
\end{equation}
where $CE(\cdot;\cdot)$ denotes the cross-entropy loss.


\begin{figure*}[tb]
	\begin{center}
		
		\centerline{\includegraphics[width=1.0\linewidth]{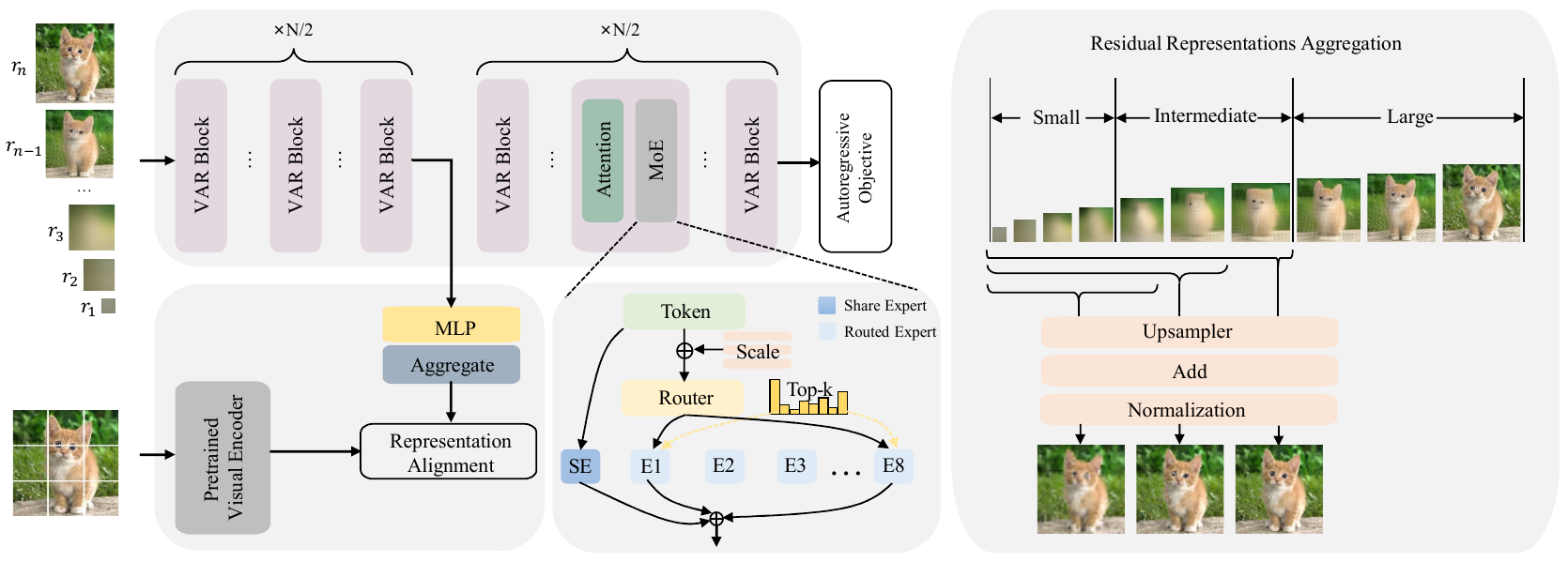}}
		\caption{\textbf{An overview of our proposed MEPA framework.} We follow the next-scale prediction paradigm of VAR~\cite{VAR}, and introduce MoE layers to route tokens across scales, allowing them to adaptively select experts. During training, the outputs of the middle block in small and intermediate scales are aggregated to align with a pretrained visual encoder's features. } 
		\vspace{-1.0em}
		\label{framework}
	\end{center}
\end{figure*}

\section{Method}





\subsection{Overview of MEPA}

We introduce MEPA, a scale-aware MoE framework for guiding efficient representation learning with semantically-enriched visual representations, as shown in Fig.~\ref{framework}. MEPA extends the model to a multi-experts architecture adapting to specialization at every scale, as detailed in Sec.~\ref{moe}. It decouples model capacity across scales and induce scale-specialized representation learning. Besides, MEPA heuristically aggregates residual latents for further alignment, as detailed in Sec.~\ref{alignment}. This allows the VAR model to leverage semantically-rich external representations for generation, preventing semantic errors at early scales. Our key contributions lie in systematically evaluating different MoE architectures and aggregating strategies for multi-scale residual representations from the perspectives of semantic level, training efficiency, and generation performance.

\subsection{Scale-Aware Token-Routed Mixture of Experts}
\label{moe}

\textbf{Mixture of Experts.} There is a fundamental insight of MoE that different components of a model can specialize in handling distinct tasks. By dynamically activating only relevant modules, MoE enables efficient scaling of model capacity while maintaining computational efficiency. Let's formally define a general MoE layer with $M$ experts, which are implemented as Feed-Forward Network (FFN) with the same architecture, denoted by $E_1, E_2,...,E_M$. The MoE layer employs a router to select a part of experts to process input tokens and replaces vanilla FFN layer in the transformer block. Consider input $x \in \mathbb{R}^{B \times N \times d}$ where $B$ is the batch size, $N$ is token length of one sample, and $d$ is the hidden dimension. The output of the MoE layer is the average weighted result of multi experts, defined as follows:
\begin{equation}
    \label{moe_output}
    y_i = \sum_{j=1}^{m} G_{i,j}E_j(x_i), x_i \in \mathbb{R}^{B \times 1 \times d},
\end{equation}
where $i$ is the token indice along the sequence length, $j$ is the expert indice, $G_{i,j}$ indicates whether the $j$th expert receives the $i$th token. Obtaining $G_{i,j}$ involves two components: a router to compute the token-to-expert affinity score $S_{i,j}$ and a gating function to achieve sparse expert activation. The routing scheme is evidently crucial for tailoring the specialized learning and preference relationship of experts, which will be detailed below. Here, a common gating function is defined as
\begin{equation}
    \label{gate}
    G_{i,j} =
    \begin{cases}
    S_{i,j}, & S_{i,j} \in topK(\{S_{i,j}\}_{j=1}^{M}, m), \\
    0, & otherwise.
    \end{cases} 
\end{equation}

	

\begin{table}[tb]
	\caption{\textbf{Mean and variance of routed tokens by different MoE.} We separately train two VAR-d16 models based on SMoE and STMoE. Then we compute the average activated times of each expert across all layers. For clarity, we normalize the total number of processed tokens to 64.
	}
	\centering
	\setlength{\tabcolsep}{7pt} 
	\scalebox{1.0}{
		\begin{tabular}{cccccccccc}
			\toprule
			Expert & $E_1$ & $E_2$ & $E_3$ & $E_4$ & $E_5$ & $E_6$ & $E_7$ & $E_8$ & Variance \\
			\midrule
			SMoE & 11.35 & 5.61 & 9.13 & 5.18 & 5.71 & 9.23 & 11.32 & 6.46 & 6.62 \\
			STMoE & 7.92 & 7.64 & 11.56 & 6.91 & 6.44 & 7.54 & 8.31 & 7.67 & 2.41 \\
			\bottomrule
		\end{tabular}
	}
	\label{Mean_tokens_processed_by_expert}
\end{table}

\noindent\textbf{Scale-routed MoE.} We first design a Scale-routed MoE (SMoE) structure for specialized representation learning at different scales. A routing matrix $W \in \mathbb{R}^{d \times M}$ and learnable scale embeddings $SE \in \mathbb{R}^{K \times d}$ are utilized to calculate the scale-to-expert affinity matrix $S \in \mathbb{R}^{B \times N \times M}$:
\begin{equation}
  \label{affinity}
  S = {Softmax}_E(SE(scale(x))W), 
\end{equation}
where ${Softmax}_E$ denotes the softmax operation along the expert axis, $scale(\cdot)$ denotes finding the scale indice corresponding to the input token. SMoE framework sets up a scale-based router clearly assigning tokens at different scales to distinct experts, thereby decoupling model capabilities across scales and boosting multi-scale representation learning. 

However, SMoE can only provide the same activation scheme for all tokens at the same scale. The routing homogenization within the same scale will cause new problems. Due to the significant difference in the number of tokens across scales, simple scale-based routing leads to an imbalanced load for experts, severely affecting training efficiency. As shown in Tab.\ref{Mean_tokens_processed_by_expert}, the variance of the token numbers processed by different experts in the SMoE model are relatively large. Moreover, different tokens within the same scale may have varying semantic richness and importance due to their positions. SMoE applies an identical expert activation pattern to all tokens within the same scale, which limits the model's capability and causes routing homogenization within the same scale. Due to the above shortcomings, SMoE only achieves the suboptimal generation performance, as illustrated in Tab.\ref{tab:MoE-abl}.

\noindent\textbf{Scale-aware Token-routed MoE.} To handle these issues, we further propose a Scale-aware Token-routed MoE (STMoE) structure, which integrates the scale embeddings into tokens to perceive the scale information, and then employs a token-based router to adaptively enable specialized representation learning and enhance the flexibility of routing. STMoE calculate the token-to-expert affinity matrix $S \in \mathbb{R}^{B \times N \times M}$ as follows:
\begin{equation}
  \label{affinity_new}
  S = {Softmax}_E((SE(scale(x)) + x)W).
\end{equation}
In practice, the VAR transformer adds positional and scale embeddings to input tokens, naturally meeting the need for scale awareness. Benefiting from scale-aware tokens, STMoE can not only assign experts with distinct semantic granularity to different tokens within the same size, but also efficiently learn scale-specificity with superior expert load balancing (as demonstrated in Tab.\ref{Mean_tokens_processed_by_expert} and Tab.\ref{tab:MoE-abl}).

\noindent\textbf{Load Balance Loss.} Directly MoE strategies often encounter the issue of load imbalance. Existing study~\cite{zuo2021taming} has shown that evenly activated experts in an MoE layer can lead to better performance. Besides, expert load balancing is crucial for training efficiency in distributed training scenarios. To achieve balanced loading among different experts, we have incorporated a load-balance loss, $\mathcal{L}_b$, which is widely used in previous works~\cite{lepikhin2020gshard, fedus2022switch}.
\begin{equation}
  \label{Balance}
  \mathcal{L}_b = M \sum_{j=1}^{m} (\frac{1}{N}\sum_{i=1}^{N} \mathbb{I}(G_{i,j} \neq 0)) (\frac{1}{N}\sum_{i=1}^{N} S_{i,j}),
\end{equation}
where $\mathbb{I}(G_{i,j} \neq 0)$ denotes the indicator function that image token $i$ selects expert $j$.

\subsection{Residual Aggregation and Semantic Guidance}
\label{alignment}

To prevent semantic errors at early stages of autoregressive generation, we introduce a Semantic Guidance (SG) mechanism that distills VAR representations toward pretrained self-supervised visual features, following the existing work of representation alignment~\cite{repa,VAVAE}.

Unlike conventional alignment approaches that constrain the final representation, our design is motivated by the causal structure of VAR. Since generation proceeds from small to large scales, predictions at later stage are strictly conditioned on representations produced at earlier stage. Consequently, inaccurate semantics generated at early stage (small and intermediate scales) can propagate and be amplified in subsequent generation. Therefore, semantic regularization should be imposed at early scales rather than only at the final scale. 

However, directly aligning individual residual representations is suboptimal. VAR constructs a multi-scale residual representation space, while self-supervised vision encoders learn patch-based complete representations. There is an inherent gap between VAR and self-supervised representations leading to mismatched alignment. Moreover, pretrained vision encoders are optimized for discriminative tasks rather than generation, and their representations may not preserve fine-grained details required at later stage (large scales). To better inject semantic information into the VAR feature space, we first transform the residual representation space into progressively enriched aggregated representations before alignment.

Specifically, residual features are sorted in ascending order of scale, and then they are upsampled to match the shape of the self-supervised representations. We perform cumulative summation on the features from the first scale to each subsequent scale to obtain a set of aggregated ones with increasing semantic richness and information content. Consider an input image $x$ and its output of pretrained encoder $g \in \mathbb{R}^{N \times D}$, where $N$ is the patch numbers and $D$ is the channel dimension. Let $m_{\phi}(h_{\theta})_{\{1,2,...,K\}}$ denote a projection of the VAR transformer output $h_{\theta} = f_{\theta}(c,r_1',r_2',\ldots,r_{K-1}')$ that through a multilayer perception (MLP) $m_{\phi}$. In practice, we select the output of the intermediate layer in a VAR transformer, avoiding straightforward alignment and reaching the best generation performance. The group of aggregated representations can be formulated as
\begin{equation}
  \label{cumulative_representations}
  z_{j} = \sum_{i=1}^{j} Up(m_{\phi}(h_{\theta})_{i}, {sz}_K).
\end{equation}
Instead of aligning the fully aggregated representation, we selectively apply semantic guidance to aggregated features at small and intermediate scales. This design directly reflects the autoregressive dependency: by strengthening semantic correctness at early stage, we provide a reliable foundation for subsequent large-scale prediction, thereby reducing semantic error propagation.

The objective of semantic guidance maximizes patch-wise similarity between the pretrained representation $g \in \mathbb{R}^{N \times D}$ and the selected aggregated features:
\begin{equation}
\label{alignmentloss}
\mathcal{L}_{SG}(\theta, \phi) = - \mathbb{E}_{x} \left[ \frac{1}{N} \sum_{i=1}^{N} \sum_{j \in F} sim(g^{i}, z_{j}^{i}) \right],
\end{equation}
where $i$ denotes the patch index, $sim(\cdot,\cdot)$ is cosine similarity, and $F$ represents the selected early-stage aggregation group (i.e., aggregated features at scales 1-5, scales 1-6, and scales 1-7).

The final training objective combines semantic guidance with the original autoregressive loss:
\begin{equation}
\label{loss}
\mathcal{L} = \mathcal{L}_{SG} + \lambda \mathcal{L}_{CE},
\end{equation}
where $\lambda > 0$ controls the balance between autoregressive learning and semantic regularization.

By explicitly enhancing semantic modeling at small and intermediate scales, semantic guidance stabilizes the autoregressive generation chain and improves both convergence speed and final generation quality. 


\section{Experiments}

\subsection{Setup}

\textbf{Benchmarks and Baselines.} In this paper, we evaluate the generative capability on the ImageNet 256×256 class-conditional benchmark and compare our quality results with state-of-the-art image generation models, the metrics include Fréchet inception distance (FID), inception score (IS), precision and recall. Our baselines cover GAN, Diffusion, AR (next-token prediction paradigm), and existing VAR methods. For a fair comparison, we report the number of parameters, the number of steps required to generate an image, and the number of training epochs for each approach.

\begin{figure}[tb]
	\begin{center}
		\centerline{\includegraphics[width=0.98\linewidth]{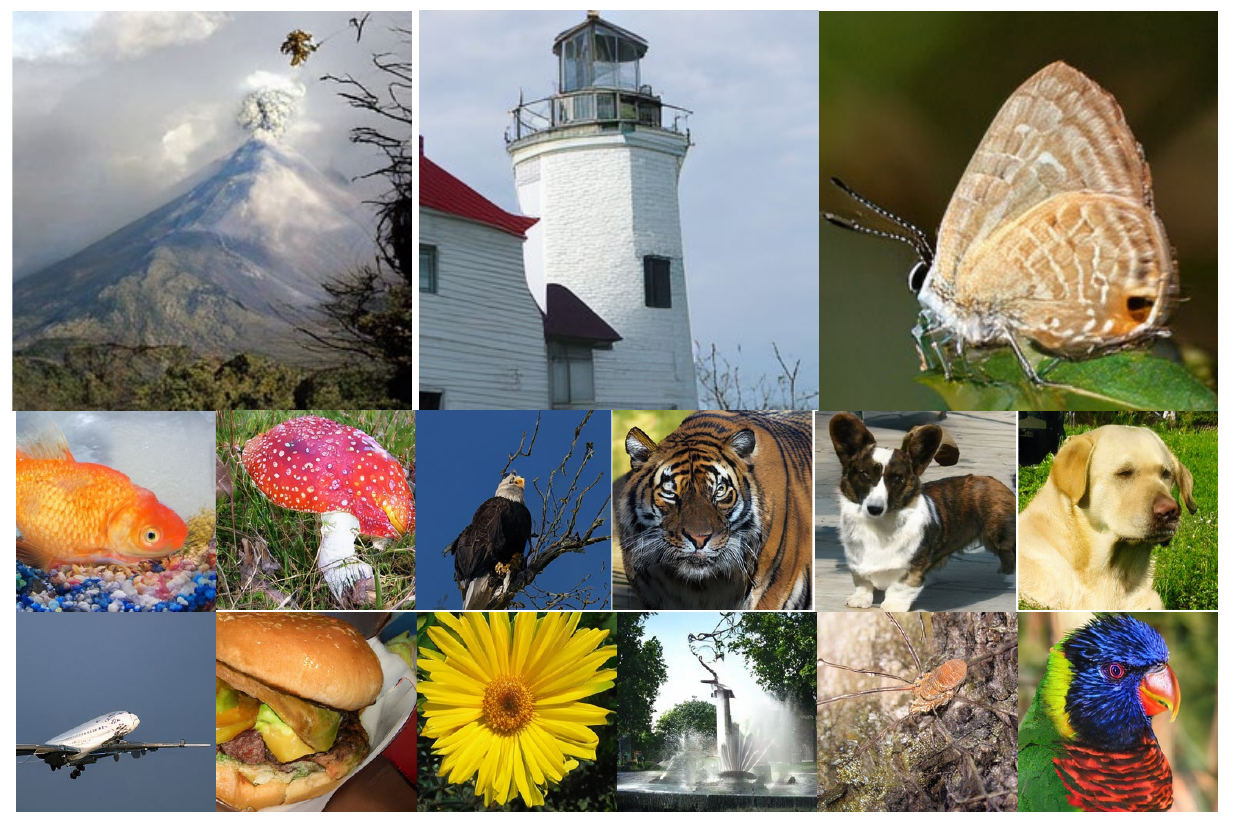}}
		\caption{\textbf{Generation results of MEPA on VAR-d16.} We use classifier-free guidance
with $w = 4.0$. } 
		\vspace{-1.0em}
		\label{pictures_of_our methods}
	\end{center}
\end{figure}

\noindent\textbf{Implementation Details.}  We train MEPA models with depths 12 and 16 on ImageNet-1K dataset~\cite{imagenet}. All models are trained under identical configurations.  We use the AdamW optimizer with a batch size of 96 and a weight decay of 0.05,  $\beta_1 = 0.9$, $\beta_2 = 0.95$. The basic learning rate is scheduled to decay from $1e-4$ to $1e-5$ following a linear annealing strategy, which is the same as VAR~\cite{VAR}. The hyper-parameter $\lambda = 0.5$.

\noindent\textbf{Comparative Method.} We compare our method with several representative approaches across GANs, diffusion models, and autoregressive models, including BigGAN~\cite{brock2018large}, GigaGAN~\cite{kang2023scaling}, StyleGAN-XL~\cite{sauer2022stylegan}, ADM~\cite{dhariwal2021diffusion}, CDM~\cite{ho2022cascaded}, LDM~\cite{rombach2022high}, DiT~\cite{peebles2023scalable}, VQGAN~\cite{vqgan}, RQTran~\cite{lee2022autoregressive}, RCG~\cite{li2023self}, and MaskGIT~\cite{chang2022maskgit}. In addition, we compare our baseline VAR with other state-of-the-art autoregressive generative models, including MAR~\cite{MAR}, FAR~\cite{FAR}, LlamaGen~\cite{llamaGen}, PAR~\cite{PAR}, FlexVAR~\cite{flexvar}, and SpectralAR~\cite{spectralar}. Compared with these methods, we emphasize that our approach is designed to achieve high training efficiency rather than state-of-the-art generative performance.

\begin{table*}[!htbp]
	\caption{\textbf{Qualitative comparison on class-conditional ImageNet 256×256}. $\uparrow$ and $\downarrow$ denote that higher and lower values are better, respectively. Param represents the number of parameters, while Step indicates the number of steps required to generate an image. $\ddag$ means we reproduce this baseline with our setting.}
	\label{tab:results}
	\centering
	\setlength{\tabcolsep}{4pt} 
	\renewcommand{\arraystretch}{1.0}
		\begin{tabular}{c|l|cccc|ccc}
			\toprule
			Type & Model  & FID$\downarrow$  & IS$\uparrow$   & Precision$\uparrow$ & Recall$\uparrow$  & Param & Step & Epoch \\
			\midrule
			    \multicolumn{8}{c}{Generative Adversarial Networks (GANs)}  \\
			GAN & BigGAN~\cite{brock2018large}         & 6.95  & 224.5 & 0.89 & 0.38             & 112M  & 1    &   -    \\
			GAN & GigaGAN~\cite{kang2023scaling}        & 3.45  & 225.5 & 0.84 & 0.61            & 569M  & 1    &    -   \\
			GAN & StyleGan-XL~\cite{sauer2022stylegan}    & 2.30  & 265.1 & 0.78 & 0.53          & 166M  & 1    &     -  \\
			\midrule
			   \multicolumn{8}{c}{Diffusion Models }    \\
			Diff & ADM~\cite{dhariwal2021diffusion}            & 10.94 & 101.0 & 0.69 & 0.63     & 554M  & 250  &    -   \\
			Diff & CDM~\cite{ho2022cascaded}            & 4.88  & 158.7 &   -   &     -          &    -   & 8100 &    -   \\
			Diff & LDM-4-G~\cite{rombach2022high}        & 3.60  & 247.7 & 0.87 & 0.48           & 400M  & 250  &    -   \\
			Diff & DiT-L/2~\cite{peebles2023scalable}        & 5.02  & 167.2 & 0.75 & 0.57       & 458M  & 250  &     -  \\
			Diff & DiT-XL/2~\cite{peebles2023scalable}       & 2.27  & 278.2 & 0.83 & 0.57       & 675M  & 250  &    -   \\
			\midrule
			   \multicolumn{8}{c}{Autoregressive Models}        \\
			AR & VQGAN-re~\cite{vqgan}       & 18.65 & 80.4  & 0.78 & 0.26             & 227M  & 256  &    -   \\
			AR & RQTran~\cite{lee2022autoregressive}.        & 13.11 & 119.3 &  -    &        -   & 821M  & 68   &    -   \\
			AR & RCG~\cite{li2023self}            & 3.49  & 215.5 &   -   &            -          & 502M  & 20   &    -   \\
			AR & MaskGIT~\cite{chang2022maskgit}        & 6.18  & 182.1 & 0.80 & 0.51            & 227M  & 8    & 300   \\
			AR & MAR-B~\cite{MAR}          & 2.31  & 281.7 & 0.82 & 0.57                         & 208M  & 64   & 800   \\
			AR & FAR-B~\cite{FAR}          & 4.26  & 248.9 & 0.79 & 0.51                         & 208M  & 10   & 400   \\
			AR & LlamaGen-L~\cite{llamaGen}     & 3.80  & 248.3 & 0.83 & 0.51                    & 343M  & 256  & 300   \\
			AR & LlamaGen-XL~\cite{llamaGen}    & 3.39  & 227.1 & 0.81 & 0.54                    & 775M  & 256  & 300   \\
			AR & PAR-L~\cite{PAR}          & 3.76  & 218.9 & 0.84 & 0.50                         & 343M  & 147  & 300   \\
			AR & PAR-XL~\cite{PAR}         & 2.61  & 259.2 & 0.82 & 0.56                         & 775M  & 147  & 300   \\
			\midrule
				\multicolumn{8}{c}{Visual Autoregressive Models}        \\
			VAR & VAR-d16~\cite{VAR}        & 3.55  & 280.4 & 0.84 & 0.51                        & 310M  & 10   & 200   \\
			VAR & VAR-d20~\cite{VAR}        & 2.95  & 302.6 & 0.83 & 0.56                        & 600M  & 10   & 250   \\
			VAR & FlexVAR-d16~\cite{flexvar}    & 3.05  & 291.3 & 0.83 & 0.52                    & 310M  & 10   & 180   \\
			VAR & FlexVAR-d20~\cite{flexvar}    & 2.41  & 299.3 & 0.85 & 0.58                    & 600M  & 10   & 250   \\
			VAR & SpectralAR-d16~\cite{spectralar} & 3.02  & 282.2 & 0.81 & 0.55        & 310M  & 64   &     200  \\
			VAR & SpectralAR-d20~\cite{spectralar} & 2.49  & 305.4 &   -   &  -          & 600M  & 64   &    250   \\
      VAR & \textbf{MEPA-d12 (Ours) } &  2.86  &   288.44   &   0.82   &    0.55  &      252M  & 10   & 200   \\
			VAR & \textbf{MEPA-d16 (Ours)}   & 2.32  & 311.26 & 0.82 & 0.57          & 585M  & 10   & 200  \\
			\midrule
			VAR & VAR-d16$\ddag$~\cite{VAR}        & 4.10 & 241.6 & 0.85 & 0.47                      & 310M  & 10   & 100   \\
			VAR & FlexVAR-d16$\ddag$~\cite{flexvar}    & 7.68  & 176.63 & 0.77 & 0.49                    & 310M  & 10   & 100   \\
			VAR & FlexVAR-d20$\ddag$~\cite{flexvar}    & 5.77  & 215.04 & 0.77 & 0.52                    & 600M  & 10   & 100   \\
			VAR & \textbf{MEPA-d12 (Ours) } &  3.27  &   260.97   &   0.81   &    0.53  &      252M  & 10   & 100   \\
			VAR & \textbf{MEPA-d16 (Ours)}       & 2.65  & 304.60 & 0.82 & 0.56          & 585M  & 10   & 100  \\
			\bottomrule
		\end{tabular}
\end{table*}

\subsection{Main Results}

We compare MEPA with existing generative methods on the ImageNet-1K benchmark~\cite{imagenet}, including GAN, diffusion models, AR models and VAR models. To ensure a fair comparison, we report the total number of parameters activated by STMOE in MEPA and compare MEPA with the dense baseline with similar parameters. As shown in Tab~\ref{tab:results}, MEPA achieves state-of-the-art performance among all methods with the default setting of training epochs, indicating great enhancement in generative quality. To show its advantages of training efficiency, we report the performance of MEPA with 100 epochs. As shown in Tab~\ref{tab:results}, MEPA-\{d12,d16\} easily outperform other VAR methods under 100 epochs, and slightly surpass VAR-\{d16,d20\} under the default training epochs by $-0.28$ and $-0.30$ FID decrease. It gains comparable FID/IS performance to other VAR methods while requiring only about a half of training epochs, revealing remarkable enhancement in training efficiency. To validate the performance under fully converged settings, we further train VAR models with MEPA for 200 epochs, matching the default VAR training schedule to ensure sufficient convergence. As shown in Tab~\ref{tab:results}, MEPA further outperforms the corresponding VAR baseline, extending the advantage on FID to $-0.69$ and $-0.63$. These empirical results demonstrate that MEPA consistently outperforms baselines both before and after full convergence, demonstrating it indeed makes progress in both generation performance and training efficiency.


\begin{table}[!htbp]
	\vspace{-0.8em}
	\begin{minipage}[t]{0.48\linewidth}
		\centering
		\caption{\textbf{Ablation Results of STMoE and semantic guidance.} We evaluate the impact of STMoE and SG on generation performance on VAR-d16 models trained by 100 epochs. }
		  \label{tab:ab-all}
		\centering
		\scalebox{0.90}{
		\begin{tabular}{@{}ccccccc@{}}
		\toprule
		STMoE   & SG   & IS$\uparrow$ & sFID$\downarrow$ & FID$\downarrow$  & Precision$\uparrow$ & Recall$\uparrow$ \\
		\midrule
		\XSolidBold     & \XSolidBold        & 241.6  & 8.75 & 4.10 & \textbf{0.85} & 0.47 \\
		\CheckmarkBold  & \XSolidBold        & 284.2  & \textbf{8.04} & 2.99 & 0.84 & 0.54  \\
		\XSolidBold     & \CheckmarkBold     & 269.0  & 8.58 & 3.59 & 0.84 & 0.50  \\
		\CheckmarkBold  & \CheckmarkBold     & \textbf{304.6}  & 8.25 & \textbf{2.65} & 0.82 & \textbf{0.56} \\
		\bottomrule
		\end{tabular}}

	\end{minipage}
	\hfill
	\begin{minipage}[t]{0.48\linewidth}
		\caption{\textbf{Ablation Results on MoE types.} We evaluate the impacts of different types of MoE on generation performance on VAR-d16 models \textbf{without semantic guidance}. }
		\label{tab:MoE-abl}
	  \centering
	  \scalebox{0.85}{
		\begin{tabular}{@{}cccccc@{}}
			\toprule
			Model          & FID $\downarrow$ & sFID $\downarrow$ & IS $\uparrow$ & Precision & Recall \\ \midrule
			 VAR             & 4.10             & 8.75              & 241.6         & \textbf{0.85} & 0.47   \\
			+MoEEC~\cite{moeec}      & 4.06             & 8.81              & 234.9         & 0.84          & 0.48   \\
			+MoE++~\cite{moe++}      & 3.39             & 8.13              & 271.7         & 0.84          & 0.51   \\
			+SMoE       & 3.11             & \textbf{7.77}     & 280.7         & 0.84          & 0.52   \\
			+STMoE         & \textbf{2.99}    & 8.04              & \textbf{284.2}& 0.84          & \textbf{0.54}   \\ \bottomrule
		\end{tabular}}
	\end{minipage}
\vspace{-0.8em}
	
\end{table}

\subsection{Ablation Studies}

In this study, we verify the proposed MEPA through detailed ablation experiments, with a focus on evaluating the effectiveness of its core mechanisms and the contribution of each component.

\noindent\textbf{STMoE and Semantic Guidance.} We evaluate the impact of STMoE and semantic guidance on generation performance. As shown in Tab.~\ref{tab:ab-all}, both STMOE and semantic guidance can achieve significant improvement on generation performance, and their combination gains the best overall performance. This indicates that the two strategies not only function independently but also complement each other in alleviating the challenges associated with multi-scale residual representation learning.

\noindent\textbf{Routing Policy of MoE.} We evaluate the impacts of different MoE strategies on the VAR model without semantic guidance. As shown in Tab.~\ref{tab:MoE-abl}, MoEEC\cite{moeec} and +MoE++~\cite{moe++} offers no appreciable benefits in practice, as they does not incorporate scale-aware design. This makes them struggle to perceive scale changes and decouple models' learning capabilities. Although SMoE best aligns with the principle that "features of different scales require independent modeling," it still performs slightly worse than STMoE. This is caused by its routing homogenization within the same scale, consistent with our analysis in Sec.~\ref{moe}.

\begin{figure*}[tb]
	\begin{center}
		\centerline{\includegraphics[width=0.98\linewidth]{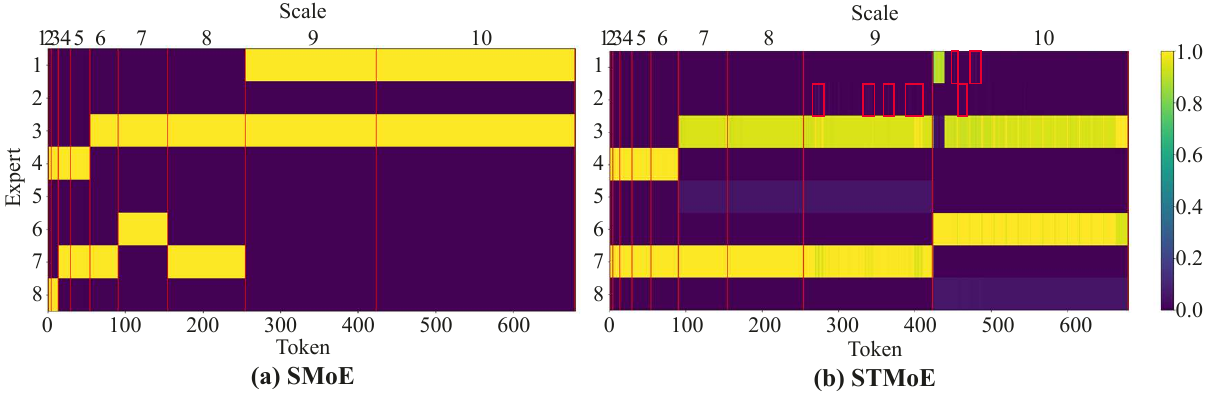}}
		\caption{\textbf{Routing heatmaps of SMoE and STMoE.} We visualize the routing heatmap of VAR-d16 models based on SMoE and STMoE in the penultimate layer. All the models are trained \textbf{without semantic guidance}. For better visualization, we highlight several points with relatively low activation values using \textcolor{red}{red boxes}.
		} 
		\vspace{-1.0em}
		\label{moe_behaviour}
	\end{center}
\end{figure*}

Besdeis, We further visualize the routing heatmaps of SMoE and STMoE in Fig.\ref{moe_behaviour}. Although both SMoE and STMoE can decouple experts' preferences across scales, STMoE exhibits two clear advantages. First, it activates eight experts, whereas SMoE utilizes only six, resulting in higher expert utilization. Second, STMoE enables adaptive routing based on token semantics and spatial positions. For example, at scale 10, Expert 3 prefers tokens at the boundary of the feature map, Expert 6 favors non-boundary tokens, and Expert 1 primarily attends to tokens in the first row of the feature map. Such a complex routing pattern emerges from adaptive learning conditioned on token-level semantic and positional information. By avoiding homogeneous routing within the same scale, STMoE provides greater flexibility and more fine-grained expert specialization, which promotes decoupled representation learning in the multi-scale feature space.

\noindent\textbf{Target Representation for Semantic Guidance.} We employ different pretrained self-supervised visual encoders for semantic guidance, as listed in Tab.\ref{tab:repa-abl1}. Specifically, we align the features from the intermediate layer in the VAR model with those extracted from pre-trained encoders. Our experimental results demonstrate that MoCov3~\cite{moco} provides no substantial benefit for promoting model convergence, while DINOv3~\cite{dinov3} achieves slightly superior performance compared to DINOv2~\cite{dinov2}. This is because the training pipeline of DINOv3 is more complex, and its prediction of structure and semantics is more accurate. Therefore, we adopt DINOv3 for semantic guidance in MEPA.


\noindent\textbf{Residual Representations Aggregation Policy.} Following the division strategy illustrated in Fig.\ref{framework}, we aggregate residual features across different scales in the VAR model and then align these aggregated features with those extracted by DINOv3. The corresponding results are reported in Tab.\ref{tab:repa-abl2}. Notably, the naive feature alignment strategy (denoted as “+SG-Last”), which applies alignment only after aggregating residual features from all scales into a complete representation, yields limited performance gains for the VAR model. We attribute this to the causal nature of the autoregressive generation process in VAR. Since generation proceeds in a coarse-to-fine manner, predictions at larger scales are conditioned on feature maps produced at smaller scales. Directly constraining the complete feature representation does not explicitly strengthen the semantic correctness of early-scale predictions, and therefore cannot effectively prevent semantic errors from propagating to later scales.


\begin{figure*}[t]
	\begin{center}
		\centerline{\includegraphics[width=0.98\linewidth]{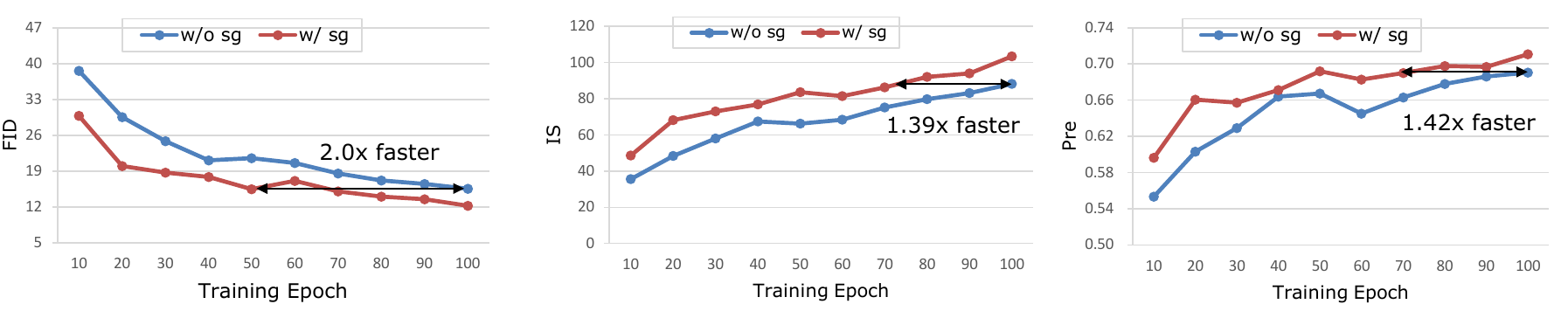}}
		\caption{\textbf{Comparison of VAR with/without semantic guidance.} We conduct a comprehensive evaluation of generation performance for models trained with and without semantic guidance. All the models are dense transformers \textbf{without MoE layers}. } 
		\vspace{-1.0em}
		\label{sg_compare}
	\end{center}
\end{figure*}

\begin{table}[!htbp]
	\begin{minipage}[t]{0.48\linewidth}
		\caption{\textbf{Ablation Results on External Self-supervised Encoders for Semantic Guidance.} We evaluate the impact of different self-supervised vision models on generation performance on VAR-d16 models trained with semantic guidance. "+SG-Mocov3", "+SG-DINOv2", and "+SG-DINOv3" denote aligning with external representations from Mocov3, DINOv2, and DINOv3, respectively. All the models are dense transformers \textbf{without MoE layers}. }
		\label{tab:repa-abl1}
	  \centering
	  \scalebox{0.8}{
		\begin{tabular}{@{}cccccc@{}}
			\toprule
			Model         		& FID $\downarrow$  & sFID $\downarrow$     & IS $\uparrow$     & Precision 		 & Recall \\ \midrule
			 VAR                    & 4.10 				& 8.75  				& 241.6 			& \textbf{0.85}      & 0.47   \\
			 +SG-Mocov3~\cite{moco}   			& 4.11 				& 8.76  				& 250.4 			& \textbf{0.85}      & 0.46   \\
			 +SG-DINOv2~\cite{dinov2}  			& 3.81 				& 8.43  				& 260.8 			& \textbf{0.85}      & 0.48   \\
			 +SG-DINOv3~\cite{dinov3}  				&  \textbf{3.59}    & \textbf{8.58}     & \textbf{269.0} & 0.84          & \textbf{0.49}  \\
			 \bottomrule
		\end{tabular}}

	\end{minipage}
	\hfill
	\begin{minipage}[t]{0.48\linewidth}
		\caption{\textbf{Ablation Results on Residual Representation Aggregation.} We evaluate generation performance of VAR-d16 models under different residual representation aggregation strategies trained with semantic guidance. "+SG-Last", "+SG-L", "+SG-S", and "+SG-M" denote selecting the last/top-3 largest/top-3 smallest/3 intermediate accumulated features, respectively. All the models are dense transformers \textbf{without MoE layers}. }
		\label{tab:repa-abl2}
	  \centering
	  \scalebox{0.9}{
		\begin{tabular}{@{}cccccc@{}}
			\toprule
			Model        & FID $\downarrow$ & sFID $\downarrow$ & IS $\uparrow$  & Precision     & Recall        \\ \midrule
			 VAR         & 4.10             & 8.75              & 241.6          & \textbf{0.85} & 0.47          \\
			+SG-Last  & 3.81             & 8.43              & 260.8          & \textbf{0.85} & 0.48          \\
			+SG-L     & 3.68             & 8.71              & 252.2          & 0.84          & 0.49          \\
			+SG-S    & 3.64             & 8.77              & 254.6          & 0.84          & \textbf{0.50} \\ 
			+SG-M     & \textbf{3.59}    & \textbf{8.58}     & \textbf{269.0} & 0.84          & 0.49          \\
			\bottomrule
		\end{tabular}}
	\end{minipage}
	\vspace{-1.0em}
\end{table}

To better align with the autoregressive dependency structure, we instead apply semantic alignment to selectively aggregated features at earlier stage. As discussed in Sec.~\ref{alignment}, we set three groups of optional aggregation scales. Then we search the best scale group by generation performances. As shown in Tab.\ref{tab:repa-abl2}, the intermediate scale group (i.e., aggregated features at scales 1-5, scales 1-6, and scales 1-7) reach the best generation quality. This result is consistent with our motivation: strengthening semantic regularization at small and intermediate scales improves the quality of the coarse semantic sketches generated at early stage. Since large-scale predictions are conditioned on these representations of earlier scales, a more accurate semantic foundation reduces error propagation and facilitates more reliable fine-grained detail generation, ultimately improving overall generation quality.

To further demonstrate the effect of semantic guidance, we compare VAR-d16 models trained with and without semantic guidance across training epochs. As shown in Fig.~\ref{sg_compare}, semantic guidance accelerates convergence, achieving up to a 2× speedup when reaching the same performance level as the baseline model. This validates that explicitly enhancing early-scale semantic modeling effectively improves both representation learning and training efficiency under the autoregressive framework.


\subsection{Discussion about representation alignment}
A key distinction is that VAR introduces residual-based, multi-scale feature space, which is absent in diffusion models studied by REPresentation Alignment (REPA)~\cite{repa}. We analyze these VAR-specific challenges (feature gap between residual and complete features, selecting specific semantic granularity for alignment) and propose tailored solutions. To our knowledge, this is the first systematic study of representation alignment in the VAR paradigm, rather than a direct extension of REPA. Moreover, MEPA consists of STMoE and semantic guidance, where our contributions to analyzing and designing the MOE architecture are also critical.

As shown in the Tab.~\ref{tab:compare_with_repa}, the improvement of MEPA outperforms REPA significantly (21.36\% vs. 12.62\% gain), proving its suitability for VAR transformers. Although SiT+REPA shows lower FID, this is expected since the VAR has weaker generation quality but higher inference efficiency than SiT. MEPA significantly boosts VAR's generation quality to approach diffusion levels while retaining VAR's distinct advantage in inference speed. In terms of time cost, MEPA increases inference time by 0.7$\times$ and training time per epoch by 4.50\%. This moderate overhead is offset by faster convergence and consistent quality gains, supporting our efficiency claim.

\begin{table*}[htbp]
  \centering
  \caption{\textbf{Qualitative comparison with REPA}. We conduct a qualitative comparison between MEPA and REPA. Param represents the number of parameters. Inference denotes the inference time per image, while Train indicates the training time of one epoch.}
  \label{tab:compare_with_repa}
  \setlength{\tabcolsep}{6pt} 
  \begin{tabular}{lllllll}
  \toprule
  Model          & Param  & Epoch & IS$\uparrow$ & FID$\downarrow$  & Inference & Train\\
  \midrule
  VAR-d20        & 600M      & 250  & 302.6                  & 2.95               & 0.50s & 1h51m \\
  MEPA-d16   & 585M    & 100   & 299.8($\pmb{-0.93\%}$)     & 2.79($\pmb{-5.42\%}$)  & 0.85s & 1h56m \\
  MEPA-d16   & 585M    & 200   & 311.3(+2.88\%)     & 2.32($\pmb{-21.36\%}$)  & 0.85s & 1h56m \\
  \midrule
  SiT-XL/2       & 675M     & 1400   & 270.3                  & 2.06             & 45s & $\_$ \\
  +REPA         & 675M     & 200    & 264.0(-2.33\%)              & 1.96(-4.85\%)      & 45s & $\_$   \\
  + REPA         & 675M     & 800  & 284.0($\pmb{+5.07\%}$)           & 1.80(-12.62\%)    & 45s & $\_$   \\
  \bottomrule
  \end{tabular}
\end{table*}
\vspace{-2.0em}

\begin{table*}[htbp]
  \centering
  \caption{\textbf{Qualitative comparison on class-conditional ImageNet 512×512.}.}
  \label{tab:compare_512}
  \setlength{\tabcolsep}{4pt} 
  \begin{tabular}{llllllll}
  \toprule
  Model          &  BigGAN & ADM & DiT-XL/2  & MaskGIT & VQGAN & VAR-d36-s & MEPA-d20 \\
  \midrule
  FID$\downarrow$ &  8.43    & 23.24   & 3.04      & 7.32   & 26.52 & 2.63 & 7.29 \\
  IS$\uparrow$    & 177.9    & 101.0   & 240.8     & 156.0  & 66.8 & 303.2 & 199.2 \\
  Epoch           & $\_$        & $\_$       & $\_$         & $\_$    & $\_$    & 350   & 66 \\
  \bottomrule
  \end{tabular}
\end{table*}
\vspace{-2.0em}

\section{Limitations and Future Work}

Due to limited computational resources, we were unable to conduct additional experiments within the current timeline. We report the results of MEPA-d20 trained at a 512×512 resolution for 66 epochs in Tab.~\ref{tab:compare_512}. Even with this restricted training budget, MEPA already outperforms several existing methods despite not reaching state-of-the-art performance. It indicates the potential of our approach for further scaling up. For future work, we intend to finish the complete training pipeline under the 512×512 setting and explore novel architectures with higher input resolutions.

\section{Conclusion}

In this work, we address the inherent challenges of multi-scale representation learning in Visual AutoRegressive (VAR) modeling. We identify two key limitations: optimization conflicts caused by sharing a single architecture across scales, and semantic error propagation induced by the causal coarse-to-fine generation process. To overcome these issues, we propose \underline{m}ulti-scale r\underline{ep}resentation \underline{a}lignment (MEPA), a scale-aware token-routed MoE framework that promotes decoupled representation learning across scales and strengthens semantic information. Extensive experiments demonstrate that our method consistently improves both training efficiency and generation quality.

\section*{Acknowledgements}
This work was supported in part by the National Natural Science Foundation of China under Grants  U22A2096, in part by Scientific and Technological Innovation Teams in Shaanxi Province under Grant 2025RS-CXTD-011, in part by the Shaanxi Province Core Technology Research and Development Project under Grant 2024QY2-GJHX-11, in part by the Establishment Fund of the State Key Laboratory of Wakefullness/Sleep and Cognition(Chongqing Institute for Brain and Intelligence), in part by the Fundamental Research Funds for the Central Universities under Grant QTZX26138, in part by the Innovation Fund of Xidian University under Grant YJSJ26022.


%
%
\bibliographystyle{splncs04}
\bibliography{main}
\end{document}